\documentclass[sigconf,natbib=false]{acmart}

\AtBeginDocument{%
  }

\copyrightyear{2024}
\acmYear{2024}
\setcopyright{rightsretained}
\acmConference[IDE '24]{2024 First IDE Workshop}{April 20, 2024}{Lisbon, Portugal}
\acmBooktitle{2024 First IDE Workshop (IDE '24), April 20, 2024, Lisbon, Portugal}\acmDOI{10.1145/3643796.3648464}
\acmISBN{979-8-4007-0580-9/24/04}

\RequirePackage[
  datamodel=acmdatamodel,
  style=acmnumeric,
  ]{biblatex}

\usepackage{color}
\usepackage{listings}
\usepackage{xcolor}
\definecolor{javared}{RGB}{6,125,23} % for strings
\definecolor{javagreen}{rgb}{0.25,0.5,0.35} % comments
\definecolor{javapurple}{RGB}{0,51,179} % keywords
\definecolor{javadocblue}{rgb}{0.25,0.35,0.75} % javadoc
 
\usepackage{xcolor}
\definecolor{javared}{RGB}{6,125,23} % for strings
\definecolor{javagreen}{rgb}{0.25,0.5,0.35} % comments
\definecolor{javapurple}{RGB}{0,51,179} % keywords
\definecolor{javadocblue}{rgb}{0.25,0.35,0.75} % javadoc
\usepackage{atbegshi,ifthen,tikz}
\usepackage{multirow}

\tikzstyle{highlighter} = [
  yellow,
  line width = \baselineskip,
]

\newcounter{highlight}[page]

\AtBeginShipout{\AtBeginShipoutUpperLeft{\ifthenelse{\value{highlight} > 0}{\tikz[remember picture, overlay]{\foreach \stroke in {1,...,\arabic{highlight}} \draw[highlighter] (begin highlight \stroke) -- (end highlight \stroke);}}{}}}

\usepackage{xspace}
\newcommand{\requirement}[1]{\textbf{R#1}\xspace} % observations

\begin{document}

%%
%% The "title" command has an optional parameter,
%% allowing the author to define a "short title" to be used in page headers.
\title{Detecting Security-Relevant Methods using Multi-label Machine Learning}

%%
%% The "author" command and its associated commands are used to define
%% the authors and their affiliations.
%% Of note is the shared affiliation of the first two authors, and the
%% "authornote" and "authornotemark" commands
%% used to denote shared contribution to the research.
\author{Oshando Johnson}
\email{oshando.johnson@iem.fraunhofer.de}
\affiliation{%
  \institution{Fraunhofer IEM}
  \streetaddress{Zukunftsmeile 1}
  \city{Paderborn}
  \country{Germany}
  \postcode{33102}
}

\author{Goran Piskachev}
\email{gpiskach@amazon.de}
\affiliation{%
  \institution{Amazon Web Services}
  \streetaddress{1 Th{\o}rv{\"a}ld Circle}
  \city{Berlin}
  \country{Germany}}

\author{Ranjith Krishnamurthy}
\email{ranjith.krishnamurthy@iem.fraunhofer.de}
\affiliation{%
  \institution{Fraunhofer IEM}
  \streetaddress{Zukunftsmeile 1}
  \city{Paderborn}
  \country{Germany}
  \postcode{33102}}

\author{Eric Bodden}
\email{eric.bodden@uni-paderborn.de}
\affiliation{%
  \institution{Paderborn University and Fraunhofer IEM}
  \streetaddress{Fürstenallee 11}
  \city{Paderborn}
  \country{Germany}}

%%
%% By default, the full list of authors will be used in the page
%% headers. Often, this list is too long, and will overlap
%% other information printed in the page headers. This command allows
%% the author to define a more concise list
%% of authors' names for this purpose.
%\renewcommand{\shortauthors}{Trovato et al.}

%%
%% The abstract is a short summary of the work to be presented in the
%% article.
\begin{abstract}

 To detect security vulnerabilities, static analysis tools need to be configured with security-relevant methods. Current approaches can automatically identify such methods using binary relevance machine learning approaches. However, they ignore dependencies among security-relevant methods, over-generalize and perform poorly in practice. Additionally, users have to nevertheless manually configure static analysis tools using the detected methods. Based on feedback from users and our observations, the excessive manual steps can often be tedious, error-prone and counter-intuitive.

In this paper, we present Dev-Assist, an IntelliJ IDEA plugin that detects security-relevant methods using a multi-label machine learning approach that considers dependencies among labels. The plugin can automatically generate configurations for static analysis tools, run the static analysis, and show the results in IntelliJ IDEA. Our experiments reveal that Dev-Assist's machine learning approach has a higher F1-Measure than related approaches. Moreover, the plugin reduces and simplifies the manual effort required when configuring and using static analysis tools. 

%A demo video\footnote{TODO: YouTube Video Link} and the
\end{abstract}

%%
%% The code below is generated by the tool at http://dl.acm.org/ccs.cfm.
%% Please copy and paste the code instead of the example below.
%%



%%
%% Keywords. The author(s) should pick words that accurately describe
%% the work being presented. Separate the keywords with commas.

%\received{20 February 2007}
%\received[revised]{12 March 2009}
%\received[accepted]{5 June 2009}

%%
%% This command processes the author and affiliation and title
%% information and builds the first part of the formatted document.
\maketitle

\section{Introduction}

With the continued rise in the number of reported software vulnerabilities \cite{hackerone}  such as those in the 2023 Common Weakness Enumeration (CWE) Top 25 Most Dangerous Software Weaknesses list \cite{top25}, more and more companies resort to Static Application Security Testing (SAST) tools to detect vulnerabilities. Yet, to be able to detect security vulnerabilities effectively, experts currently need to correctly configure and adapt the SAST tools. One required configuration comprises security-relevant methods (SRM), which are critical points in a program that have an impact on the analysis~\cite{swan}. SRMs relevant for detecting taint-style vulnerabilities include sources (methods that create data that the analysis should track), sinks (methods at which the analysis might need to raise an alarm), and sanitizers (methods that perform data sanitization). 

For Java, several machine learning (ML) approaches have been proposed  to replace the time-consuming and error-prone manual approach of detecting SRMs in large codebases \cite{susi, swan, sas}. One such approach, SWAN, classifies Java methods into sources, sinks and sanitizers as well as seven\footnote{CWE78 OS Command Injection, CWE79 Cross-site Scripting, CWE89 SQL Injection, CWE306 Missing Authentication, CWE601 Open Redirect, CWE862 Missing Authorisation, and CWE863 Incorrect Authorisation} Common Weakness Enumeration types---with high precision and recall \cite{swan}. SWAN-Assist is an IntelliJ IDEA plugin built on top of SWAN that uses active machine learning to detect SRMs \cite{swan-assist}. With minimal user effort, the plugin's semi-automatic approach improves SWAN's precision. 

The fundamental machine learning problem that SWAN addresses is a multi-label learning problem in which real-world objects that have several semantic meanings \cite{zhang2018binary} are represented by a set of labels \cite{tsoumakas2007multi,zhang2013review}. Detecting security-relevant methods in Java programs is a multi-label machine learning problem in which one assigns a subset of SRM labels to each method. In the code snippet in Listing~\ref{list:sqli}, three security-relevant methods are invoked. The source method \emph{getParameter()} creates the taint, the sanitizer \emph{encodeForSQL()} sanitizes the user data, and \emph{executeQuery()} is the sink method. In this example, the source and sanitizer methods impact the content of the SQL query that the sink method executes on the database. These methods are therefore relevant for CWE89 SQL injection. A method that writes unsanitized user input to a database and also uses the unsanitized data in a redirect should be labeled as a sink both for SQL injection (CWE89) and Open Redirect (CWE601) vulnerabilities. 

\begin{lstlisting}[caption={Code snippet with a potential SQL injection from source (l.\ref{ln:sqli:source}) to sink (l.\ref{ln:sqli:sinksqli}) mitigated using a sanitizer (l.\ref{ln:sqli:sanitizer}).}, label={list:sqli}]
protected void doPost(HttpServletRequest req, ...) {
   @String@ usr = req.getParameter("ID"); @\label{ln:sqli:source}@//source
   usr = ESAPI.encoder().encodeForSQL(new MySQLCodec(), usr); @\label{ln:sqli:sanitizer}@//sanitizer
   String query = "select * from user" +
        "where username='" + usr + "'";
   Statement stmt = conn.createStatement();
   ResultSet rs = stmt.executeQuery(query); @\label{ln:sqli:sinksqli}@//sink
\end{lstlisting}

%SWAN decomposes the multi-label problem into independent single-label binary classification problems.

Although the classification of security-relevant methods is a multi-label problem, SWAN solves it with a binary relevance (BR) approach, which decomposes the machine learning problem into several independent single-label classification problems \cite{zhang2018binary}. This approach fails to exploit the correlation among labels and can therefore produce prediction models that generalize badly~\cite{binaryrel}. Additionally, in order to use SWAN-Assist's results to configure a SAST tool, users have to manually extract the SRM information and externally verify if the configuration produces the expected results. 

In this paper, we present Dev-Assist, a holistic IntelliJ IDEA plugin that uses a multi-label machine learning approach that generalizes better in assigning SRM labels to Java methods. The plugin can use labeled SRMs to automatically generate the configuration for SAST tools, integrates a  static analysis to detect security vulnerabilities, and shows the analysis results within IntelliJ. We show that the multi-label approach outperforms existing binary relevance classification approaches for detecting SRMs and reduces manual effort when using the detected security-relevant methods.

Dev-Assist's source code, documentation and datasets are available online\footnote{https://github.com/secure-software-engineering/swan/tree/master/dev-assist}. We next explain the motivation and requirements for the plugin. Section \ref{sec:approach} describes the plugin's interface, features and back-end. We evaluate our approach in Section \ref{sec:evaluation}, and subsequently discuss the limitations and conclusions of our work.

\section{Motivation and Requirements}\label{sec:requirements}

%To solve the multi-label problem of detecting SRM types for Java methods, SWAN decomposes the problem into several single label classification problems, referred to as binary relevance (BR) \cite{zhang2018binary}. However, because BR uses an independent method which does not utilize the correlations between labels, it may produce prediction models that do not generalize well in practice \cite{zhang2018binary}.   

SWAN and SWAN-Assist have two main shortcomings, the machine learning solution for the multi-label classification problem and poor usability due to excessive manual effort.

%Although the classification of security-relevant methods is a multi-label task, SWAN solves it by using the binary relevance (BR) approach, which decomposes the machine learning problem into several independent single-label classification problems \cite{zhang2018binary}. 

It is generally known that, because BR uses an independent method that does not exploit the correlations between labels, it may produce prediction models that do not generalize well in practice~\cite{zhang2018binary}. Moreover, SWAN's BR approach has a different feature representation for each SRM label and uses a two-phase model selection approach. These design and implementation choices complicate maintenance and evaluation of the approach. 

We identified the need to improve SWAN-Assist's usability and reduce the required manual effort at a Security Testing and Verification Summer School in Belgium that was attended by more than 80 PhD students and researchers \cite{summerschool}. At the summer school, we organized a hands-on workshop session focused on detecting SRMs and using them to detect security vulnerabilities in the WebGoat \cite{webgoat} project. Participants were required to use SWAN-Assist to detect and update the list of SRMs. To detect the vulnerabilities, participants had to manually create the specifications \cite{fluenttql} necessary to configure and run the open-source taint analysis tool SecuCheck \cite{secucheck}. Although many of the participants found the tools useful, one resounding feedback was that using the SRM list to create specifications to run SecuCheck, was complex, error-prone, counter-intuitive, and required too much manual work. Many participants did not complete these tasks within the allotted time. Additionally, the participants' feedback and our observations revealed numerous bugs, errors, and improvements for SWAN-Assist and SecuCheck. 

%Many participants did not complete these tasks within the allotted time and reported that some of the tasks that required too much manual effort were error-prone, time intensive, and counter-intuitive.

Given the limitations of SWAN-Assist and SWAN's machine learning approach, a new plugin that meets the following requirements is necessary:

%To support developers and security experts in their Integrated Development Environment (IDE) with the task of detecting SRMs, generating specifications and analysing their code, we require a vulnerability detection pipeline. 

\begin{itemize}
    \item \emph{\requirement{1}: The plugin shall assign SRM labels using an ML approach that considers dependencies among SRM labels}: Given that dependencies may exist among SRM categories, the model should use this information so that it can generalize better.
    \item \emph{\requirement{2}: The plugin shall use detected SRMs to automatically generate taint-flow query specifications}: Instead of requiring users to export the detected SRMs and manually create taint-flow query specifications, the plugin should use the SRM list to automatically generate specifications for SecuCheck.
    \item \emph{\requirement{3}: The plugin shall run a taint analysis and report vulnerabilities in the IDE}: Using automatically generated or external taint-flow specifications, the plugin should be able to run a taint analysis and display the analysis results in the IDE. 
\end{itemize}

%SWAN-Assist \cite{swan-assist} partially meets \requirement{1}, however, the binary relevance approach does not consider dependencies among the SRM labels. 
%\todo{I think we can drop the previous sentence. Confuses me more than it helps.}

\section{Dev-Assist Plugin: Automated Configuration for Static Analysis}\label{sec:approach}

\begin{figure}[ht]
	\centering
	\includegraphics[scale=1.1, page=2]{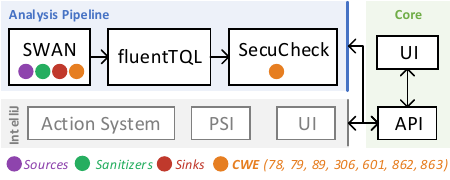}
	\caption{Dev-Assist's architecture showing the interaction of the core module, analysis pipeline, and IntelliJ platform.}
	\label{fig:dev-assist-architecture}
\end{figure}

To address the limitations of SWAN and SWAN-Assist, we present a holistic plugin, Dev-Assist, that automatically detects SRMs using a multi-label approach, generates taint-flow specifications, and also runs a static code analysis. Figure \ref{fig:dev-assist-architecture} shows the architecture of the Dev-Assist plugin which consists of a core module and analysis pipeline that are built on top of the IntelliJ platform. 

The core module contains the plugin's user interface (UI) and an application programming interface (API) that coordinates the communication with the other modules. The API interacts with the analysis pipeline by configuring, executing, and processing its results.  The API also interacts with the IntelliJ platform \cite{intellij-platform} and uses the Program Structure Interface (PSI)~\cite{psi}, which provides a syntactic and semantic code model, the Action System \cite{actionsystem} to add items to IntelliJ's menus and toolbars, and the IntelliJ IDEA UI.

The plugin was developed using the Gradle IntelliJ Plugin (1.16.1) and Java 17. In line with IntelliJ's recommendation~\cite{support} to support the latest major release and at least two previous releases, the plugin supports IntelliJ Platform version 2022.1 to 2023.3. 

%We next discuss the core and analysis pipeline modules in detail.

\subsection{Plugin Interface and Features}

%To support the new analysis pipeline and enhance the plugin's user interface, we had to implement several changes. Figure \ref{fig:dev-assist} shows the updated plugin tool window (1), dialog for updating SRMs (2), and sample vulnerability report (3).

\begin{figure*}
	\includegraphics[width=1\textwidth]{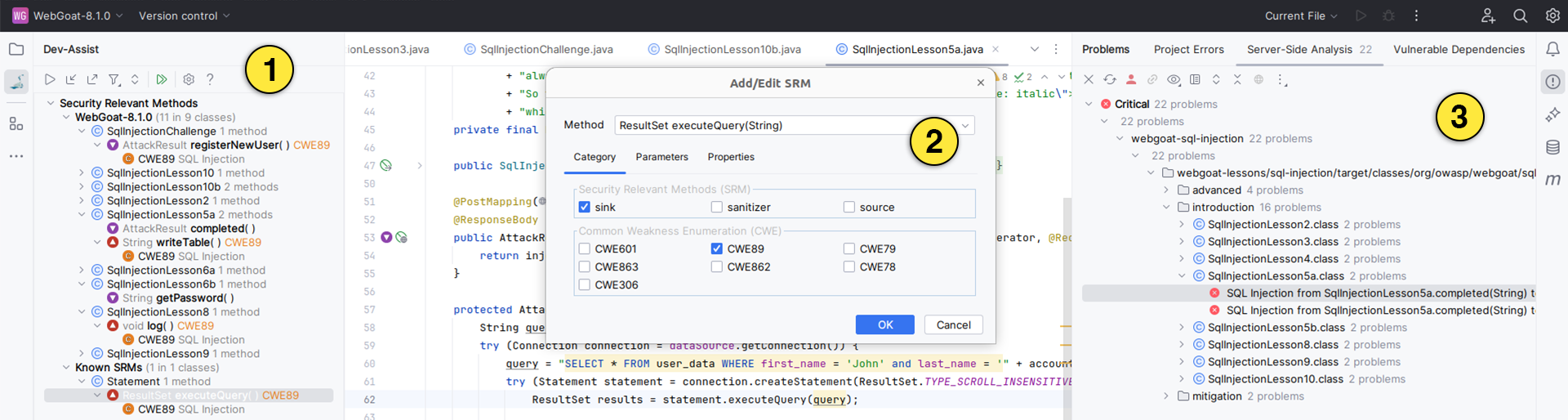}\centering
	\caption{IntelliJ window screenshot showing the Dev-Assist plugin tool window (1), method dialog (2), and analysis results (3). }
	\label{fig:dev-assist}
\end{figure*}

To support the new analysis pipeline and enhance the plugin's user interface, we had to implement several back-end and front-end changes. Figure \ref{fig:dev-assist} shows Dev-Assist's tool window (1), dialog for updating SRMs (2), and a sample vulnerability report (3).

\subsubsection{Tool Window (1)} The toolbar in the plugin's tool window contains buttons to run the analysis modules, as well as buttons to filter, expand/collapse the expandable list, and to also import/export the SRM list. The expandable list view below the toolbar shows the SRM list for methods detected in the active project as well as SRMs present in SWAN's training dataset. We have updated the expandable list to provide more information to the user without having to expand all nodes in the list.  

We added a new action to the plugin's toolbar to run SecuCheck with the generated specifications, and another action to manage the plugin settings. Previously, a user would always see a dialog when running the analysis. Now we show the dialog only when the user runs the analysis the first time and, if they wish, they can update the settings using the new toolbar action. 

\subsubsection{Method Dialog (2)} When a user selects a method in the list view or editor, the edit method dialog appears. Using this dialog, a user can view and update the method's classification, taint-flow information as well as other properties. 

Instead of two dialogs to update the classification of the SRMs and method properties, we have merged the dialogs into one dialog that uses a tabbed pane. We have replaced the double list interface for updating SRM labels with checkboxes to reduce the effort necessary to change the categories. A new tab with UI components to update the data flow properties of method parameters is also now present. The two properties, \textit{data-in}, and \textit{data-out}, are particularly important when generating taint-flow specifications as they indicate which method parameters are included in the data flow operations.

%\subsubsection{Editor Markers} In the gutter of the editor (3), the plugin annotates code by using line markers to indicate the SRMs. Upon clicking the line markers, the edit dialog appears to provide more information, and tooltips are also shown when one hovers over the markers. 

\subsubsection{Analysis Results (3)} Instead of building a new interface to show the analysis pipeline results, we opted to use the existing \textit{Problems} tool window, which displays problems detected by IntelliJ IDEA using several tabs \cite{problems}. The \textit{Server-Side Analysis} tab of the \textit{Problems} tool window displays problems detected with the code quality platform Qodana \cite{qodana} and also provides navigation to the code fragments that contain the reported problems. Using Qodana, the Static Analysis Results Interchange Format (SARIF) \cite{sarif} analysis results from Dev-Assist can be opened as local reports.

%In this paper, we target the IntelliJ Platform \cite{intellij-platform} as it offers numerous features and Application Programming Interfaces (API) for building code analysis tools. 

\subsection{AI Supported Analysis Pipeline}\label{vul-pipeline}

%To meet the requirements outlined in Section \ref{sec:requirements}, we implemented a multi-label machine learning version of SWAN and integrated SecuCheck such that detected SRMs can be used to generate \textit{fluent}TQL specifications, that SecuCheck uses to identify security vulnerabilities. The analysis pipeline module is integrated into the plugin's core module as a Gradle dependency. 

The analysis pipeline detects security-relevant methods using the multi-label implementation of SWAN, automatically generates \textit{fluent}TQL specifications and finds security vulnerability using SecuCheck. The analysis pipeline module is integrated into the plugin's core module as a Gradle dependency. Next, the three components of the analysis pipeline are discussed.

%The Dev-Assist IntelliJ plugin implements the vulnerability detection pipeline discussed in Section \ref{vul-pipeline} that automatically detects SRMs in an active project, allows users to update the SRM list, generates taint-flow specifications, and also runs the code analysis, all within the IDE. 

\subsubsection{Multi-Label SRM Detection}

%The classification of security relevant methods is a multi-label task that SWAN solves by using the binary relevance (BR) approach which decomposes the machine learning problem into several independent single label classification problems \cite{zhang2018binary}. However, because BR uses an independent method which does not utilize the correlations between labels, it may produce prediction models that do not generalize well in practice \cite{zhang2018binary}. 

To address SWAN's binary relevance limitations and fulfill \requirement{1}, we extended SWAN to use MEKA (v1.9.7) \cite{meka}, a multi-label extension of the ML Java Library WEKA\cite{weka}. The framework has been designed specifically for the implications of multi-label approaches such as modeling dependencies between variables and also the evaluation of such models which involves scoring multiple target classifications for each instance~\cite{meka}. 

\begin{figure}[ht]
	\centering
	\includegraphics[scale=0.92, page=1]{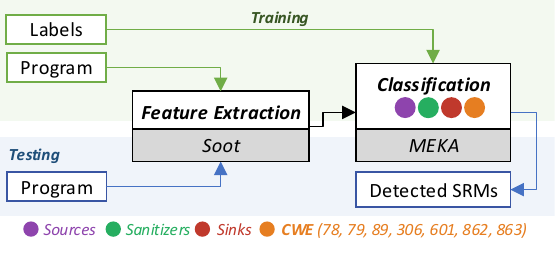}
	\caption{Diagram of Dev-Assist's multi-label machine learning approach for detecting security-relevant methods. }
	\label{fig:dev-assist-approach}
\end{figure}

The multi-label ML approach is shown in Figure~\ref{fig:dev-assist-approach}. In the training phase, we use the static program analysis framework Soot~\cite{cetus11soot} to extract information from the programs in the training dataset to create the feature vector. The feature vector and labeled methods are provided as inputs to MEKA to train and identify the best model. In the testing phase, the feature vector is provided to the trained model and the detected SRMs are exported to a JSON file. 

We updated SWAN's original dataset, which contained 235 labeled Java methods, with 1417 new methods that we extracted from public SRM lists~\cite{findsecbugs2017,sampaio2014methods, van2008owasp} and their respective Java archive (JAR) files. The updated dataset now contains 1,625 labeled methods that are stored in a JSON file in the project repository. 

%Dev-Assist reuses all of the features from SWAN, however, we have consolidated the features so that they can be used for multi-label learning. 

Instead of generating 10 feature representations, one for each of the SRM types, for SWAN's two-phase approach, all features have been merged into one feature representation. Dev-Assist reuses all of the features from SWAN, however, we have consolidated the features so that they can be used for multi-label learning. The majority of the features use a bag-of-words~\cite{bow} approach to evaluate the presence of pre-defined tokens in the signatures of methods defined or invoked in a program. Some examples of these tokens are: "database", "delete", "replac", "encod", "redirect", and others that can be used to infer what a method or class does. The remaining features are categorical features that evaluate various properties such as the modifiers and types of classes and methods.

We have implemented 12 new features that count various properties of the methods and classes after processing the programs with Soot. For a given method, we count its code lines, methods invoked, branching statements, iteration statements, exception handlers, parameters, and variables defined/used. For the class that the method belongs to, we count its code lines, functions, and tokens. With the consolidation of the existing features and addition of the new features, Dev-assist's feature vector contains 119 features (13 numeric, 99 binary and 7 categorical).
 
In the training phase, the feature extraction module uses the labels and feature vector to create a dataset that can be used to train the model. MEKA uses the Attribute-Relation File Format (ARFF) to specify the labels as multiple binary attributes and the list of instances \cite{meka}. Dev-Assist's training ARFF file contains 129 attributes (10 SRM categories and 186 features) and 1625 instances. 

Using an ARFF file that contains the training data, we can run experiments to identify a machine learning model using MEKA's command line or graphical user interface. However, instead of manually specifying a list of machine learning algorithms for our experiments, as was done for SWAN, we opted to use automated machine learning (AutoML). This approach uses a larger search space of machine learning algorithm combinations in order to identify the multi-label classifier as well as its hyperparameters \cite{wever2021automl}. 

We use the AutoML framework ML\textsuperscript{2}-Plan (v0.2.7) to recursively select, combine, and parameterize MEKA’s multi-label algorithms in order to identify the machine learning pipeline for predicting SRMs \cite{ml2plan, wever2021automl}. We configure ML\textsuperscript{2}-Plan to use 8 CPUs, 1 hour timeout, 70:30 train-test split, and the default settings recommended in the Git repository\footnote{https://github.com/starlibs/AILibs/tree/master}. The experiments were executed on a Lenovo T14 Gen 1 20UD (Ryzen 7 Pro 4750U, 1.7 GHz) laptop. ML\textsuperscript{2}-Plan selected a multi-label machine learning model using the Ensembles of Pruned Set \cite{4781214} method with logistic model trees \cite{landwehr2005logistic} as the base classifier.

%We also improved SWAN's architecture to make it more modular and maintainable, added new options to the command line interface (CLI), and fixed several known errors and bugs. These changes contribute to an enhanced user experience and easier integration into the analysis pipeline. 

\subsubsection{Automatically Generated Specifications}

Many widely used SAST tools use taint analysis to detect security vulnerabilities. SecuCheck~\cite{secucheck} is one such tool that provides a configurable analysis and can be seamlessly integrated into various Integrated Development Environments (IDEs). With SecuCheck, users can create their own taint-flow specifications using the Java internal Domain Specific Language (DSL) \textit{fluent}TQL~\cite{secucheck, fluenttql}. To create the taint-flow specifications with \textit{fluent}TQL, users must explicitly and manually specify SRMs such as source, sink, sanitizer, and propagator as well as information from the method signatures. For example, the user would need to enter the fully qualified class names for the SRMs as well as other properties necessary for tracking data flow. 

The current approach of exporting the SRM list and requiring a user to manually create the taint-flow specifications is a tedious and error-prone process. To improve this process and meet \requirement{2}, we integrate the \textit{fluent}TQL API into the analysis pipeline to generate the taint-flow specifications from a list of SRMs. The approach provides a central way to manage the information necessary for creating the specifications, namely the SRM list which can be stored as a JSON file. The taint-flow specifications can then be used to configure SecuCheck in order to detect security vulnerabilities.

%Identifying and specifying SRMs manually is relatively easy for smaller Application Programming Interfaces (APIs), it becomes a laborious and error-prone task for larger APIs. To assist developers, security experts specify such SRMs in the existing tools such as SecuCheck~\cite{secucheck} and FindSecBugs~\cite{findsecbugs}. However, these pre-defined lists of SRMs only cover popular APIs and therefore these lists are incomplete. 

%To enhance the process of identifying and specifying SRMs, we integrate SWAN to automatically detect SRMs specific to the code being analyzed. From the list of detected SRMs, we automatically generate taint-flow specifications for SecuCheck analysis using APIs the \textit{fluent}TQL DSL. 

\subsubsection{Detecting Vulnerabilities with SecuCheck}

To run SecuCheck \cite{secucheck}, the path of the project to be analyzed, taint-flow specifications, and other analysis options can be provided via its command line interface. The analysis results are exported to a SARIF \cite{sarif} file.

%The next component in the vulnerability detection pipeline, is the taint analysis which analyses a project and report potential CWEs that were detected. In our pipeline, we use the previously discussed tool SecuCheck. Using SecuCheck's command line interface, we provide the path of the project to be analyzed, security specifications and other command line options to configure the analysis. 

%https://www.jetbrains.com/help/qodana/qodana-ide-plugin.html#ide-plugin-local-report
To fulfill \requirement{3}, we integrated SecuCheck in the analysis pipeline so that we can use the generated taint-flow specifications described in the previous section. To make this possible, we refactored the tool's implementation and architecture to expose integration points. We also updated SecuCheck to ensure that it correctly follows the SARIF 2.1.0 \cite{sarif} specification and generates a valid SARIF file. 

\section{Evaluation}\label{sec:evaluation}

Dev-Assist's main motivations were to improve the machine learning approach for detecting security-relevant methods and to reduce the manual effort when using SRMs to configure static analysis tools. We assess these contributions in the sections that follow.

\subsection{Multi-label Machine Learning Approach}

We compare Dev-Assist's multi-label approach with SWAN-Assist's binary relevance approach using the cross-validation results and also their performance on a real-world project.

To assess the in-sample performance, we did a 10-fold cross-validation with a 70:30 train-test split of the new training dataset for both approaches. Table \ref{tab:in-sample} shows the average precision, recall, and F1-score (harmonic mean of the precision and recall) over the cross-validation folds. The F1-Score is often used to estimate the performance of machine learning models, therefore, a higher F1-Score indicates better performance. Using SWAN's F1-scores as the baseline, we see that Dev-Assist outperforms SWAN-Assist for 9 SRM labels (sinks, sources, CWE78, CWE79, CWE89, CWE306, CWE601, CWE862, and CWE863). Moreover, for 3 of these SRMs (CWE78, CWE79 and CWE601), Dev-Assist's F1-Scores are 1.5-2 times higher than those of SWAN which were all lower than 0.35. For sanitizers, SWAN-Assist (0.89) performs minimally better than Dev-Assist (0.87). The low performance for CWE78 (0.55) and CWE602 (0.37) can be attributed to the imbalanced training dataset, where these CWEs have the lowest number of examples compared to the other CWE labels. Imbalanced datasets make it difficult for ML models to identify the minority class which results in lower scores \cite{krawczyk2016learning}.

%Despite being trained with more examples, SWAN-Assist did not perform that much better in comparison to its original dataset. This could probably be impacted by a combination of the feature and model selection steps used by SWAN-Assist.  %by the i because of its binary relevance approach that does not seem to generalize well.

%5 minutes --- The chosen classifier is: #27: meka.classifiers.multilabel.meta.BaggingML- [-S, 1, -I, 10, -P, 67, -W, meka.classifiers.multilabel.CC, --, -S, 0, -W, weka.classifiers.functions.SimpleLogistic, --, -I, 0, -M, 500, -H, 50, -W, 0.0]
\newcommand{\STAB}[1]{\begin{tabular}{@{}c@{}}#1\end{tabular}}
\addtolength{\tabcolsep}{-1.6pt}   
\begin{table}[t]
\begin{tabular}{cccccccccccc}
\toprule
% \multicolumn{11}{c}{\textbf{\textit{Dev-Assist}}} \\ \hline

& &  &  &  & \multicolumn{7}{c}{\textbf{Common Weakness Enumeration}} \\ 
  \cmidrule(lr){6-12}
 &  & \textbf{Sa} & \textbf{Si} & \textbf{So} & \textbf{78}  &\textbf{79} & \textbf{89} & \textbf{306} & \textbf{601} & \textbf{862} & \textbf{863}\\

%old data
%\multirow[c]{3}{*}{\STAB{\rotatebox[origin=c]{90}{\textit{Dev}}}} &\textbf{P} &0.82 & 0.84 & 0.66 & 0.91 & 0.75 & 0.53 & 0.90 & 0.95 & 0.74 & 0.84\\
%&\textbf{R} &0.80 & 0.86 & 0.49 & 0.85 & 0.69 & 0.29 & 0.85 & 0.92 & 0.65 & 0.84\\
%&\textbf{F} &0.81 & 0.85 & 0.56 & 0.88 & 0.71 & 0.37 & 0.87 & 0.93 & 0.69 & 0.84 \\ \hline

\multirow[c]{3}{*}{\STAB{\rotatebox[origin=c]{90}{\textit{Dev}}}} &\textbf{P} & 0.90 & 0.75 & 0.70 & 0.75 & 0.60 & 0.88 & 0.91 & 0.44 & 0.84 & 0.84 \\
&\textbf{R} & 0.85 & 0.80 & 0.63 & 0.43 & 0.57 & 0.75 & 0.81 & 0.32 & 0.78 & 0.79 \\
&\textbf{F} & 0.87 & 0.78 & 0.66 & 0.55 & 0.58 & 0.81 & 0.86 & 0.37 & 0.81 & 0.81 \\ \hline

%\multicolumn{11}{c}{\textbf{\textit{SWAN-Assist}}} \\ \hline
\multirow[c]{3}{*}{\STAB{\rotatebox[origin=c]{90}{\textit{SWAN}}}} &\textbf{P} &0.89 & 0.79 & 0.66 & 0.87 & 0.41 & 0.86 & 0.87 & 0.41 & 0.89 & 0.83\\
&\textbf{R} &0.90 & 0.72 & 0.57 & 0.22 & 0.24 & 0.68 & 0.75 & 0.18 & 0.65 & 0.66\\
&\textbf{F} &0.89 & 0.75 & 0.61 & 0.34 & 0.30 & 0.75 & 0.81 & 0.24 & 0.74 & 0.73 \\
\bottomrule
\end{tabular}
\caption{Dev-Assist (Dev) and SWAN-Assist (SWAN) 10-fold cross-validation precision (P), recall (R), and F1-Score (F) results for sanitizers (Sa), sinks (Si), sources (So), and CWEs. }
\label{tab:in-sample}
\vspace{-12pt}
\end{table}
\addtolength{\tabcolsep}{1.6pt}

To evaluate how well Dev-Assist's generalizes on real-world projects, out-of-sample, we used Dev-Assist to detect SRMs in the Android 13 (API level 33) \cite{android13} project. Then, we randomly selected 20 methods predicted for each SRM label and manually verified the predictions. Dev-Assist's precision for the selected methods is shown in Table \ref{tab:out-sample}. The average precision was 0.72 and CWE862/CWE863 have the highest precision (0.95). CWE89 had the lowest precision (0.35) and is consistent with the cross-validation results. Despite low precision on the random samples for CWE89, Dev-Assist correctly labeled 74 sinks and CWE89 methods in the \textit{android.database.sqlite.SQLiteDatabase} class. In comparison to SWAN-Assist, more than 80\% of these methods are not classified as security-relevant which indicates that Dev-Assist is more precise in practice.

\addtolength{\tabcolsep}{-1.6pt}   
\begin{table}[t]
\begin{tabular}{cccccccccccc}
\toprule
% \multicolumn{11}{c}{\textbf{\textit{Dev-Assist}}} \\ \hline

 &  &  &  & \multicolumn{7}{c}{\textbf{Common Weakness Enumeration}} & \\ 
  \cmidrule(lr){5-11}
   & \textbf{Sa} & \textbf{Si} & \textbf{So} & \textbf{78}  &\textbf{79} & \textbf{89} & \textbf{306} & \textbf{601} & \textbf{862} & \textbf{863} & \textbf{$\overline{P}$}\\

\textbf{P} &0.9 & 0.65 & 0.85 & 0.75 & 0.5 & 0.35 & 0.6 & 0.6 & 0.95 & 0.95 & 0.72 \\
\bottomrule
\end{tabular}
\caption{Dev-Assist's Precision (P) on 200 randomly selected methods from the Android 13 library.}
\label{tab:out-sample}
\vspace{-20pt}
\end{table}
\addtolength{\tabcolsep}{1.6pt}

\subsection{Security Vulnerability Specifications and Detection }

%In this section, we evalute the \rk{this sentence is incomplete}

To reduce the manual effort required to use SRM lists exported by SWAN-Assist to create security specifications that can be used by taint analysis tools to detect security vulnerabilities, several changes were made to Dev-Assist. We will assess the new approach by evaluating the number of steps and the average time required to accomplish the same task using SWAN-Assist and Dev-Assist.

For our evaluation, we use the tasks that the participants of the Summer School mentioned in Section \ref{sec:requirements} were required to perform in order to use SWAN-Assist and SecuCheck to find vulnerabilities in the WebGoat project. The tasks are listed in Table \ref{tab:workflow} along with the average duration for each task that were estimated by two of the workshop organizers that supported the participants during the hands-on session. The duration only considers the time when the user actually performs an activity and therefore excludes the run and/or wait time. 

\begin{table}[t]
\begin{tabular}{c|p{5.6cm}|p{1.1cm}}
\toprule
% \multicolumn{11}{c}{\textbf{\textit{Dev-Assist}}} \\ \hline

\textbf{No.} & \textbf{Task} & \textbf{Duration}   \\ 
1 & Run analysis to detect SRMs & 0.5 Min. \\
2 & Update detected SRMs and add new ones & 5 Min. \\
3 & Export SRMs to JSON file & 0.5 Min. \\
4 & Create \textit{fluent}TQL specifications using SRM information (for e.g., method signature) in the JSON file & 7 Min. \\
5 & Compile and export \textit{fluent}TQL specifications & 0.5 Min. \\
6 & Configure SecuCheck with the exported specifications and set analysis options & 4 Min. \\
7 & Run SecuCheck and export analysis results to a SARIF file & 0.5 Min.\\
%8 & Open SARIF file with an external tool & 1 \\
8 & Interpret taint analysis results and validate them in the source code & 7 Min. \\
\hline 
& & 25 Min \\

\bottomrule

\end{tabular}
\caption{Workflow showing the steps and average duration when using SWAN to manually configure SecuCheck.}
\label{tab:workflow}
\vspace{-20pt}
\end{table}

Based on Table \ref{tab:workflow}, the participants spent 76\% of their time using the exported SRMs to create the security specifications (Tasks 4-5), set up and run SecuCheck (Tasks 6-7), and to interpret the results (Task 8). Dev-Assist consolidates and automates five tasks (Task 3-7) such that with the click of a button all of these steps will be performed and the analysis results will be available in Qodana. This reduces the number of tasks and the estimated time to perform the tasks by 50\%. Consolidating these steps also means that a user can obtain results quicker without the need for much manual intervention.

%Steps 4 and 6 were particularly problematic in the workshop as many participants were not able to create the specifications and run the analysis in the allotted time. 

%The manual effort to copy the method signatures from the JSON file (Task 4) and to format it according to the requirements of \textit{fluent}TQL was error-prone and counter-intuitive for many participants. Similar difficulties were observed and shared by participants when configuring SecuCheck  

%Moreover, the configuration of SecuCheck was also not very clear for some participants and they also had difficulties interpreting the results in the SARIF file exported by SecuCheck as they had to use text editors and then switch to the IDE to verify the results. Due to these challenges, we had to provide extra time.

Given that the results are visible within the IDE and intuitively shown, the effort to interpret and action the results (Task 7) is also reduced. A developer needs not to switch to an external tool or manually search for the results within the SARIF file. 

%Dev-Assist therefore reduces the number of steps from 8 to 4: run the analysis, update the SRM list, run the taint analysis and then action the results. 

%Figure~\ref{fig:workflow} depicts the 9 steps in the workflow that uses the results from SWAN-Assist to detect security vulnerabilities. 

\section{Limitations and Threats to Validity}
Despite our efforts to avoid known pitfalls \cite{dosanddonts} when using ML for computer security, there are some limitations in our approach. The training examples are collected from major Java libraries, which have higher code quality in comparison to real-world projects. As in all related work, most features in our approach use information from method signatures, therefore, poorly named classes and methods pose a problem. Additionally, our training data is imbalanced and is not representative, as the proportion of SRM and non-SRM methods does not conform to the real world.

%\section{Related Work}
%META-DES experiments with alternative machine learning algorithms but \cite{android-srm}

%SuSi~\cite{susi} is an ML approach that uses binary classification to label methods in the Android framework into sources, sinks and Android-specific categories. SWAN~\cite{swan} generalizes SRM detection for Java applications and extends SuSi to detect sanitizers, authentication methods and seven CWE vulnerability types. Sas et al.~\cite{sas}  extended SuSi, introduced the need for generalizing the detection of SRMs for general Java libraries, and motivate the need for detecting CWE labels.  All of these approaches use a binary relevance approach to detect SRMs, whereas Dev-Assist uses a multi-label approach. 

\section{Conclusion}

In this paper, we present a multi-label machine learning approach for detecting security-relevant methods in Java programs using the Dev-Assist IntelliJ IDEA plugin. The plugin's ML approach considers correlation among SRM types, automatically generates specifications, and also runs a code analysis. We described Dev-Assist's features, interface, architecture, and its underlying multi-label machine learning model. Using the plugin, developers and SAST experts can create and verify SRM lists, which are required to correctly configure static analysis tools, with less manual effort and more precision.

\begin{acks}
We thank Rohith Shanmuganathan for helping with the implementation of the analysis pipeline and IntelliJ plugin. We acknowledge support by the German Federal Ministry of Education and Research (BMBF) under grant number 16KIS1257 (AI-DevAssist).

\end{acks}

%%
%% Print the bibliography
%%
\printbibliography

\end{document}